\documentclass{bmvc2k}

\newif\ifsubmit

\submittrue

\definecolor{brown}{rgb}{.5,.27,.07}

\ifsubmit
\newcommand{\cwwu}[1]{}

\newcommand{\ctliu}[1]{}
\newcommand{\revctliu}[1]{#1}
\else
\newcommand{\cwwu}[1]{{\bf \textcolor{blue}{C.-W. Wu: #1}}}

\newcommand{\ctliu}[1]{{\bf \textcolor{brown}{C.-T. Liu: #1}}}
\newcommand{\revctliu}[1]{\textcolor{brown}{#1}}
\fi

\title{Spatially and Temporally Efficient Non-local Attention Network for Video-based Person Re-Identification}

\addauthorthree{Chih-Ting Liu}{jackieliu@media.ee.ntu.edu.tw}{1}{2}{3}
\addauthorthree{Chih-Wei Wu}{cwwu@media.ee.ntu.edu.tw}{1}{2}{3}
\addauthor{Yu-Chiang Frank Wang}{ycwang@ntu.edu.tw}{2}
\addauthorthree{Shao-Yi Chien}{sychien@ntu.edu.tw}{1}{2}{3}

\addinstitution{
IOX Center\\
National Taiwan University, Taiwan
}
\addinstitution{
Department of Electrical Engineering\\
National Taiwan University, Taiwan
}
\addinstitution{
Graduate Institute of Electronics Engineering\\
National Taiwan University, Taiwan
}

\runninghead{LIU ET AL.}{ST-Efficient Non-Local Network for Video-based Person Re-ID}

\usepackage{times}
\usepackage{enumitem}
\usepackage{multirow}
\usepackage{amssymb}
\usepackage[export]{adjustbox}
\usepackage{graphicx}
\usepackage{tabularx}
\usepackage{algorithmic}

\def\etal{\emph{et al}\bmvaOneDot}
\newcommand{\mycaption}[2]{\caption{\textbf{#1.}~#2}}
\begin{document}

\maketitle

\begin{abstract}
Video-based person re-identification (Re-ID) aims at matching video sequences of pedestrians across non-overlapping cameras. 
%
It is a practical yet challenging task of how to embed spatial and temporal information of a video into its feature representation.
While most existing methods learn the video characteristics by aggregating image-wise features and designing attention mechanisms in Neural Networks, they only explore the correlation between frames at high-level features.
%
In this work, we target at refining the intermediate features as well as high-level features with non-local attention operations and make two contributions. 
(i) We propose a Non-local Video Attention Network (NVAN) to incorporate video characteristics into the representation at multiple feature levels.
%
(ii) We further introduce a Spatially and Temporally Efficient Non-local Video Attention Network (STE-NVAN) to reduce the computation complexity by exploring spatial and temporal redundancy presented in pedestrian videos. 
%
%
%
%
Extensive experiments show that our NVAN outperforms state-of-the-arts by $\mathbf{3.8\%}$ in rank-1 accuracy on MARS dataset and confirms our STE-NVAN displays a much superior computation footprint compared to existing methods. Codes are available at \url{https://github.com/jackie840129/STE-NVAN}.

\end{abstract}
\vspace{-3mm}

\section{Introduction}
\label{sec:intro}

Person re-identification (Re-ID) tackles the problem of retrieving pedestrian images/videos across non-overlapping cameras. 
%
Previous approaches mostly focus on image-based Re-ID, where each pedestrian possesses multiple images for retrieval~\cite{market,trip,cuhk03,viper,dukefeature,msmt17,grid,caviar}.
Recently, video-based Re-ID has drawn significant attention in literature since retrieving pedestrian videos is more realistic and critical in real-world surveillance applications~\cite{prid2011,ilidsvid,mars,dukevideo}.
%
With the emergence of large-scale video-based Re-ID datasets~\cite{mars,dukevideo}, researchers design Deep Neural Networks to learn robust representation for videos~\cite{mars,rcnn,forest,diversity,snip,sta}.

To perform video-based Re-ID, typical methods require learning a mapping function to project the video sequences to a low-dimensional feature space, where Re-ID can then be performed by comparing distances between samples.
%
As demonstrated by numerous works, training the convolutional Neural Network (CNN) as a mapping function has dominated over classic methods with hand-crafted features~\cite{sdalf,lomo,kissme}.
%
%
Usually, they obtain features for a sequence by aggregating image features with average or maximum pooling~\cite{rcnn,mars}.
%
However, their approaches fail to handle occlusion or spatial misalignment in video sequences since it treats all images in a sequence with equal importance~\cite{snip}.
%
%
%
%
In order to distill relevant information for Re-ID, \revctliu{some works integrate Recurrent Neural Network to learn the spatial-temporal dependency in an end-to-end training manner~\cite{rcnn,recurrent,deepfusion}.} Recently, several works propose attention mechanism to weight the importance of different frames or different spatial locations to aggregate a better representation~\cite{diversity,snip,sta}.
%
While these methods successfully capture both the spatial and temporal characteristics of video sequences, they only explore the aggregation of high-level features for representation, which might not be sufficiently robust for fine-grained classification tasks such as Re-ID~\cite{HACNN,yu2018hierarchical,shih2017deep}.
%

In this paper, we first aim to improve the representation for video sequences by exploiting spatial and temporal characteristics in both low-level and high-level features.
%
Inspired by Wang~\etal~\cite{non-local}, we propose a Non-local Video Attention Network (NVAN) by introducing the non-local attention layer into an image classification CNN model.
%
The non-local attention layer enriches the local image feature with global sequence information by generating attention masks according to features of different frames and different spatial locations.
%
By inserting non-local attention layers at different feature levels, NVAN explores the spatial and temporal diversity of a sequence and alters its feature representation subsequently rather than combining individual image features with a set of weights as in previous works.
%
Our NVAN model surpasses all state-of-the-art video-based Re-ID methods by a large margin on the challenging MARS~\cite{mars} dataset, proving that exploiting global information for multi-level features is crucial for learning representation for video sequences.
%

While applying non-local attention layer to multi-level features significantly improves the Re-ID performance, it comes at a great cost in terms of computation complexity.
%
In fact, it increases the total floating point operations (FLOP) by $99.3\%$, making it difficult to scale up to practical applications.
To alleviate such challenge, we take advantage of the space-time redundancy in pedestrian videos and propose a Spatially and Temporally Efficient Non-local Video Attention Network (STE-NVAN).
%
We first reduce the granularity of attention masks in non-local attention layers by exploiting the spatial redundancy exhibited in pedestrian images.
%
On the other hand, we explore the temporal redundancy between video frames to aggregate image-wise information into a representative video feature with a hierarchical structure.
By reducing the computation complexity both spatially and temporally, our STE-NVAN cut down $72.7\%$ of FLOP compared to original NVAN with only $1.1\%$ drop in rank-1 accuracy on MARS dataset.
%
Our proposed STE-NVAN demonstrates a much superior trade-off between performance and complexity compared to existing video-based Re-ID methods.
%
The contribution of our work can be summarized as follows:
\begin{itemize}[noitemsep]
%
\item We introduce the non-local attention operation into the backbone CNN at multiple feature levels to incorporate both spatial and temporal characteristics of pedestrian videos into the representation.
%
\item We significantly reduce the computation count for our Non-local Video Attention Network by exploring the spatial and temporal redundancy presented in pedestrian videos.
%
\item Extensive experiments validate that our proposed model not only outperforms state-of-the-art methods in Re-ID accuracy but also requires less computation count than existing attention methods for video-based Re-ID.
\end{itemize}
\section{Related Work}

In this section, we briefly review the related works regarding image-based person Re-ID, video-based person Re-ID and the usage of attention mechanisms for the Re-ID problem.

Image-based person Re-ID has been extensively studied over the years. 
With the success of CNNs~\cite{domaindrop,svdnet,IDE,trip,HACNN}, deep features learned from the networks has replaced hand-crafted features~\cite{viper,market,caviar,lomo} for representing pedestrian images.
As suggested by Zheng~\etal~\cite{ppp}, these networks can be categorized into discriminative learning and metric learning.
%
Discriminative learning learns deep features for identity classification with the help of the cross-entropy loss~\cite{domaindrop,svdnet,IDE}.
%
As for metric learning, Hermans~\etal~\cite{trip} use the triplet loss to teach the network to push together features of the same person and pull away features of different people.
In this work, we utilize both loss functions to train our network for video-based person Re-ID.

Video-based person Re-ID is an extension of image-based person Re-ID. 
%
Zheng~\etal~\cite{mars} introduce a large-scale dataset to enable the learning of deep features for video-based Re-ID.
They first train a CNN to extract image features then aggregate them into a sequence features with average/maximum pooling. 
Other works~\cite{rcnn,forest,recurrent} adopt Recurrent Neural Networks to summarize image-wise features into a \revctliu{single feature} by exploiting temporal relation within a sequence.

Recently, attention mechanisms are introduced for capturing spatial and temporal characteristics of pedestrian sequences within the deep features. 
%
Xu~\etal~\cite{jointatten} introduce the joint attentive spatial and temporal pooling network to extract sequence features by jointly considering the query and gallery pairs \revctliu{with an affinity matrix}. %
Li~\etal~\cite{diversity} learn attention weights to combine features of different spatial locations and different temporal frames into a sequence feature.
%
Chen~\etal~\cite{snip} utilize techniques in~\cite{attention_is_all} to perform self-attention on each video snippet and co-attention between video snippets for learning sequence features.
%
Fu~\etal~\cite{sta} learn sequence features by mining features of discriminative regions and select important frames with a parameter-free attention scheme.
%
While these works achieve promising results by introducing spatial and temporal attention on top of high-level features obtained from image-based CNNs, they overlook the importance of utilizing video characteristics at intermediate feature levels.
In contrast, our proposed NVAN is able to refine intermediate features with spatial and temporal information of videos and our efficient STE-NVAN model substantially reduces the computation cost for incorporating video characteristics at lower feature levels. 
\vspace{-3mm}
\section{Proposed Method}

Given an image sequence of any pedestrians, we aim to learn a CNN to extract its feature \revctliu{representation} that enables video-based person Re-ID in the embedding space.
The key to learning a representative feature for a sequence is to incorporate video characteristics into the feature itself.
To this end, we introduce the non-local attention layer into the CNN to explore the spatial and temporal dependency of a video sequence.
We propose a Non-local Video Attention Network (NVAN) in Sec.~\ref{sub:NVAN} to apply such operations at different feature levels.
However, we observe incredibly large computation complexity with the introduction of attention mechanisms.
Hence, we further propose the Spatially and Temporally Efficient Non-local Video Attention Network (STE-NVAN) in Sec.~\ref{sub:STEN} to alleviate the computation cost by exploiting spatial and temporal redundancy which exists in pedestrian videos.

%
%
%
%

\subsection{Non-local Video Attention Network}
\label{sub:NVAN}

%
%

\begin{figure*}[t!]
	\centering
    \includegraphics[width=\textwidth]{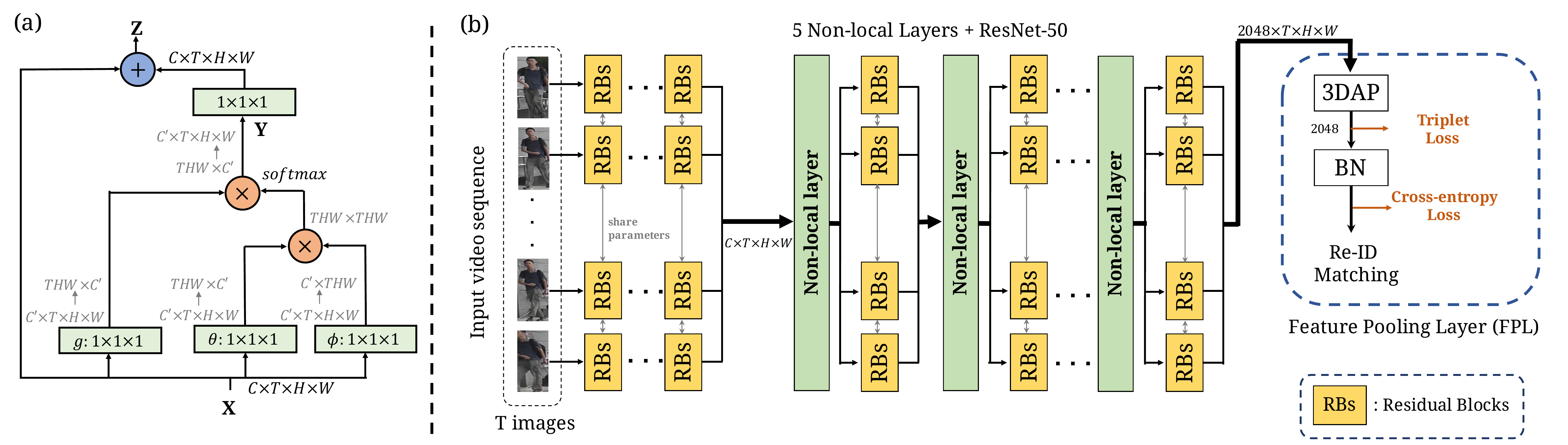}
    \mycaption{(a) Details of Non-local Attention Layer. (b) Overview of our Non-local Video Attention Network (NVAN)}{In NVAN, given $T$ sampled images as input, the 5-Non-local ResNet-50 network generates $T$ features, which incorporates the spatial and temporal information of videos at multi-levels with the help of Non-local Attention Layers. The features are then pooled into one vector in FPL for loss optimization and Re-ID matching.}
    \label{fig:NVAN}
    \vspace{-3mm}
\end{figure*}

To extract features for an image sequence, we take input as a subset of video frames selected by restricted random sampling (RRS) strategy and forward through a backbone CNN network incorporating non-local attention layers and a feature pooling layer (FPL) to obtain the representation vector for video-based Re-ID, as shown in Figure~\ref{fig:NVAN} (b).
\vspace{-3mm}

\paragraph{Restricted Random Sampling (RRS).}
There are several ways to handle the long-range temporal structure.
%
%
To balance speed and accuracy, we adopt the restricted random sampling strategy~\cite{diversity,wang2016temporal}. 
Given an input video $\mathbf{V}$, we divide it into $T$ chunks $\left\{C_{t}\right\}_{t=[1,T]}$ of equal duration. 
For training, we randomly sample an image $I_t$ in each chunk.
As for testing, we use the first image of each chunk. 
The video is then represented by the ordered set of sampled frames $\left\{I_{t}\right\}_{t=[1,T]}$.

%
%

\paragraph{Non-local Attention Layer.} 
To embed video characteristics into the features, we introduce the non-local layer proposed by Wang~\etal~\cite{non-local} into the backbone CNN, as illustrated in Figure~\ref{fig:NVAN}~(a).
%
%
Given an input feature tensor $X \in \mathbb{R}^{C \times T \times H \times W}$ obtained from a sequence of $T$ feature maps of size $C\times H\times W$, we desire to exchange information between features across all spatial locations and frames.
Let $x_i \in \mathbb{R}^C$ sampled from $X$, the corresponding output $y_i \in \mathbb{R}^C$ of non-local operation can be formulated as follow:
\begin{equation}
\label{eq:non-local}
y_i = \frac{1}{\sum_{\forall j}{e^{\theta(x_i)^T\phi(x_j)}}}\sum_{ \forall j}{e^{\theta(x_i)^T\phi(x_j)}g(x_j)}.
\end{equation}
%
Here, $i,j=[1,THW]$ indexes all locations across a feature map and all frames.
%
%
%
We first project $x$ to a lower dimensional embedding space $\mathbb{R}^{C^{\prime}}$ by using linear transformation functions $\theta, \phi, g$ ($1\times 1\times 1$ convolution).
Then, the response of each location $x_i$ is computed by the weighted average of all positions $x_j$ by using Embedded Gaussian instantiation. The Equation~\ref{eq:non-local} in non-local layer is a self-attention mechanism which is also mentioned in~\cite{non-local}.
%
The overall non-local layer is finally formulated as $Z=W_zY+X$, where the output of non-local operation is added to the original feature tensor $X$ with a \revctliu{transformation $W_z$ ($1\times 1\times 1$ convolution) that maps $Y$ to the original feature space $\mathbb{R}^{C}$}.
%
%
%
%
%
%
%
The intuition behind the non-local operation is that when extracting features at a specific location in a specific time, the network should consider the spatial and temporal dependency within a sequence by attending on the non-local context.
In our person Re-ID scheme, we embed five non-local layers into our backbone \revctliu{CNN which is a ResNet-50 network~\cite{resnet}} to comprehend the semantic relation presented in videos, as shown in Figure~\ref{fig:NVAN}~(b).
%
%
%

\paragraph{Feature Pooling Layer (FPL).}
%
After passing the image sequence through the backbone CNN and non-local attention layers, we employ the feature pooling layer to obtain the final feature for Re-ID, shown in Figure~\ref{fig:NVAN}~(b).
We apply 3D average pooling (3DAP) along the spatial and temporal dimension to aggregate the output features of each image into a representative vector, followed by a batch normalization (BN) layer~\cite{bn}. 
%
%
We train the network by jointly optimizing the cross-entropy loss and the soft-margin batch-hard triplet loss~\cite{trip}.
%
%
%
%
Interestingly, we empirically find that optimizing cross-entropy loss on the final feature while optimizing triplet loss on the feature before BN results in the best Re-ID performance.
%
%
A rational explanation is that the embedding space without normalization is more suitable for distance metric learning such as the triplet loss, while the normalized feature space forces the model to classify samples on a more constraint angular space with cross-entropy loss~\cite{trip,sphereface,arcface,cosface}.

\subsection{Spatially and Temporally Efficient Non-local Video Attention Network}
\label{sub:STEN}

While our proposed NVAN is able to capture sophisticated properties of video sequence with the help of non-local operations, we observe a significant increase in the computation complexity as shown in Table~\ref{tab:ablation1}, \revctliu{where FLOP ramp up from $30.4$G to $60.0$G.}
For scaling NVAN to practical usage scenarios, we introduce two complexity reduction techniques to cut down the computation count.
\vspace{-3mm}

%
%
%
%
\begin{figure*}[t!]
	\centering
    \includegraphics[width=\textwidth]{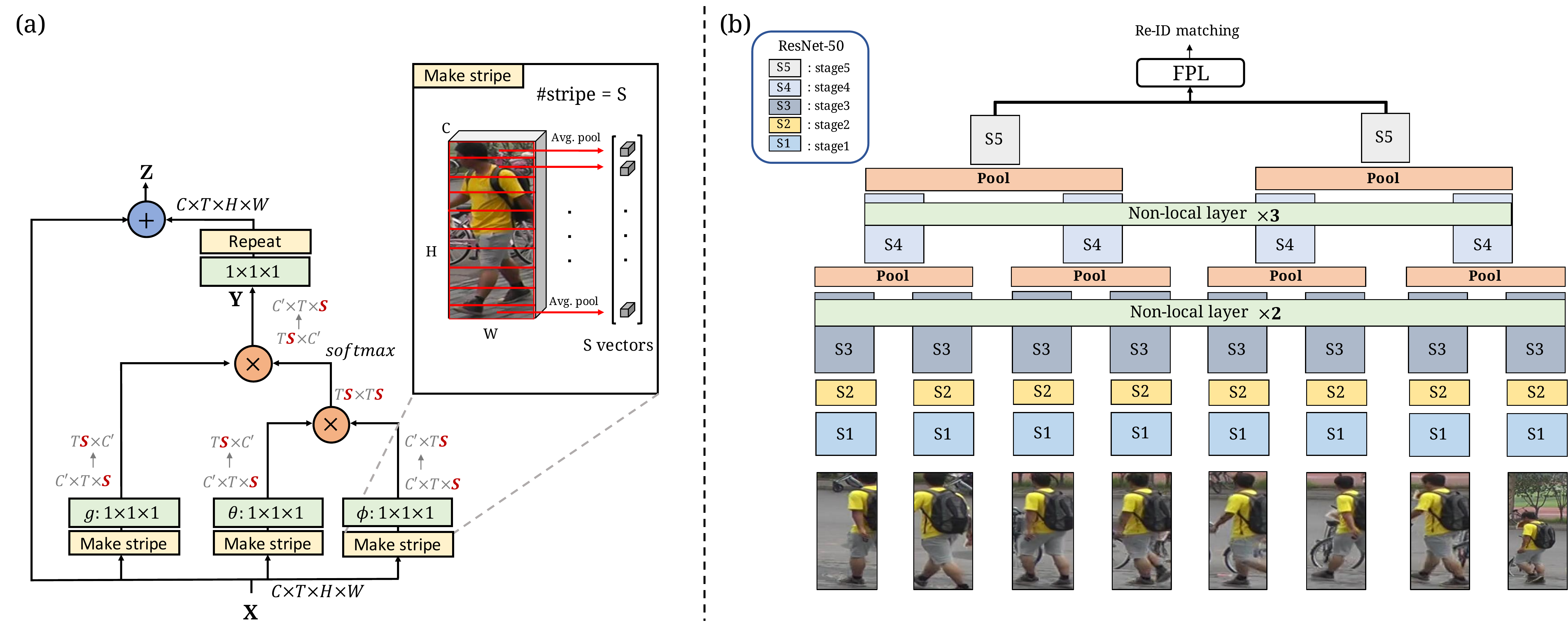}
    \mycaption{(a) Spatial Reduction Non-local Layer. (b) Temporal Reduction with Hierarchical Structure}{Details are explained in Section~\ref{sub:STEN}. Noting that, for figure (a), before the residual addition, we repeat the tensor of shape $C\times T\times S$ to $C\times T\times H\times W$. As for figure~(b), we apply max-pooling across adjacent features after the stages with non-local layers to construct our hierarchical structure.}
    \label{fig:all}
    \vspace{-3mm}
\end{figure*}
\paragraph{Spatial Reduction with Pedestrian Part Characteristics.}
Originally, the introduced non-local operations perform dense affinity calculation between features of all $THW$ positions to obtain a fine attention mask.
This results in heavy computation of complexity $O(C^\prime T^2 H^2 W^2 + CC^\prime THW)$ for each non-local attention layer. 
Applying the \revctliu{non-local attention} layer to lower feature levels incurs larger complexity since low level features are typically of higher $H,W$.
To alleviate such effect, we group the features along the horizontal direction to form a more compact representation of the feature tensor.
The intuition is that pixels of the same horizontal stripe tend to share similar characteristics which can be utilized to generate coarse representation of the image.
It is worth noting that while similar ideas have been explored in Re-ID literature~\cite{pcb,deepfusion,cheng2016person,lomo}, they use this concept to generate finer features for Re-ID.
In contrast, we exploit this redundancy to obtain coarser representation.
We partition the original feature tensor $X \in \mathbb{R}^{C \times T \times H \times W}$ into $S$ horizontal groups by adding the ``Make stripe'' module at the input of non-local operations.
The resulting tensor $X^\prime \in \mathbb{R}^{C \times T \times S}$ requires only $O(C^\prime T^2 S^2 + CC^\prime TS)$ to complete the operation, which is irrelevant to the spatial size of feature maps.
This dramatically reduces the computation complexity and enables us to deploy non-local operation to lower feature levels with constant computation cost.
We name it Spatial Reduction Non-local Layer and illustrate the idea in Figure~\ref{fig:all}~(a).

\vspace{-3mm}
\paragraph{Temporal Reduction with Hierarchical Structure.}
During our experiments, we observe that features refined by non-local operations are often temporally similar as non-local operation aims to embed global temporal information into the features.
%
%
Inspired by this observation, we exploit the temporal redundancy between features of different frames and propose a hierarchical structure to reduce the heavy computation of extracting sequence feature.
We illustrate this idea in Figure~\ref{fig:all}.
%
%
After passing a sequence of images through a series of convolutions (Residual blocks) and non-local attention layers, we apply max pooling across features of adjacent frames and reduce the temporal feature dimension by a factor of 2.
%
%
We perform the same reduction operation after another stacks of Residual blocks until the temporal dimension is reduced to 2, which is then sent to FPL for final feature summarization.
This temporal reduction technique cuts down the computation required for extracting sequence feature with Residual blocks and non-local attention layers. 
%
By applying both the Spatial Reduction Non-local Layers and the Hierarchical Temporal Reduction structure, we come up with the final Spatially and Temporally Efficient Non-local Video Attention Network (STE-NVAN) for video-based person Re-ID.
%

\section{Experiments}
\label{sec:exp}

We evaluate our approach on two large-scale video-based person Re-ID datasets, MARS~\cite{mars} and DukeMTMC-VideoReID~\cite{dukevideo}. 
We conduct ablation studies to validate the effectiveness of non-local operations and the two proposed reduction methods. 
We compare our NVAN and STE-NVAN models to existing state-of-the-arts to demonstrate that our proposed models display superior performance while requiring less computation counts.
\vspace{-3mm}
\subsection{Experimental Setup}
\paragraph{Datasets and Evaluation Protocal.} MARS~\cite{mars} is one of the large video-based person Re-ID datasets, consisting of 17,503 tracks and 1,261 identities.  
Each track has 59 frames on average.
Deformable Part Model~\cite{dpm} is employed to detect pedestrians and GMCP~\cite{gmcp} is used to track pedestrians.
%
To make the dataset even more challenging, they include 3,248 distractor tracks in the dataset.
DukeMTMC-VideoReID~\cite{dukevideo} is another large-scale benchmark recently introduced for video-based person Re-ID. %
%
It comprises 4,832 tracks and 1,404 identities and 408 distractor identities.
%
Each track contains 168 frames on average. 
Detection and tracking ground truth are manually labeled.
%
In the following literature, DukeMTMC-VideoReID will be abbreviated as ``DukeV'' for convenience. 
In our experiments, we adopt the standard train/test split and report both rank-1 accuracy (R1) and Mean Average Precision (mAP) to evaluate the Re-ID performance. 

\vspace{-3mm}
\paragraph{Implementation Detail.}
For the RRS strategy described in Sec.~\ref{sub:NVAN}, we segment each video into $T=8$ chunks and sampled $8$ images as the input sequence. 
Each frame is resized to $256\times 128$ and synchronously augmented with random horizontal flip for each track. 
We adopt the ImageNet pre-trained ResNet-50~\cite{resnet} as our backbone network, and modified $conv5\_1$ to stride 1 instead of stride 2 to better adapt the Re-ID task.
%
%
For our NVAN, we insert 2 non-local attention layers after $conv3\_3, con3\_4$ and another 3 after $con4\_4, con4\_5, con4\_6$ respectively. 
As for STE-NVAN, we set $S=16$ in Spatial Reduction Non-local layer and perform max-pooling right after the second and the fifth non-local attention layer to reduce temporal dimension.
%
%
We train our network for 200 epoch with both cross-entropy loss and triplet loss~\cite{trip} and choose Adam~\cite{adam} optimizer with an initial learning rate of $10^{-4}$ and decay it by 10 every $50$ epochs.
Following the suggestion in~\cite{trip}, we sample 8 identities, each with 4 tracks, to form a batch of size $8\times 4 \times 8=256$ images.
%
%
%
%

\vspace{-3mm}
\begin{table}[t]
    \centering
    \mycaption{Comparisons of different baselines with two reduction methods}{This table shows the performance results and the computation count of baseline models, NVAN and STE-NVAN. The ``Reduc.'' is the abbreviation of Reduction.}
    \label{tab:ablation1}
    \begin{tabular}{l|c|cc|cc|c}
    \hline
    \multirow{2}{*}{Method} & Feature &\multicolumn{2}{c|}{MARS} & \multicolumn{2}{c|}{DukeV}& \multirow{2}{*}{\# FLOP} \\ 
    \cline{3-6} 
      & Aggregation & R1         & mAP       & R1    & mAP    &     \\ 
     \hline \hline
     ResNet-50   & FPL & 87.3       & 79.1      & 95.0  & 92.7   &  30.4~G \\
    ResNet-50   & max-FPL  & 86.3 & 76.6      & 95.4  & 92.4 & 30.4~G\\
    \hline
    NVAN   & FPL & \textbf{90.0}       & \textbf{82.8}      & \textbf{96.3}  & \textbf{94.9} & 60.0~G  \\
    NVAN+Spatial Reduc.   & FPL & 89.7      & 82.5      & 96.3  & 94.7    & 30.4~G \\
    NVAN+Temporal Reduc.   & FPL & 89.2       & 81.2      & 95.6  & 93.7    & 40.4~G \\
    \hline
    STE-NVAN   & FPL & 88.9      &   81.2    & 95.2  & 93.5   & \textbf{16.5~G} \\
    \hline
    \end{tabular}
    \vspace{-1mm}
\end{table}

\subsection{Ablation Studies}
\paragraph{Effectiveness of Non-local Attention Layer and Two Reduction Methods.}
We first compare our NVAN model with two baseline models to demonstrate the power of non-local operations. 
The two baseline models (ResNet-50) use the same backbone network as NVAN but without non-local attention layers.
The only difference between the two baselines is that one replace the 3DAP in FPL with 3D maximum pooling operation.
%
%
The first three rows in Table~\ref{tab:ablation1} illustrate the results. 
It reveals that non-local operations improve the R1 and mAP significantly by $2.7\%,3.7\%$ on MARS and $1.3\%,1.6\%$ on DukeV.
The improvement confirms the effectiveness of incorporating spatial and temporal characteristics in the sequence feature of different semantic levels.
%
%
%
However, we observe an dramatic $99.3\%$ increase in FLOP accompanying the introduction of non-local operations.
%
%
Therefore, we propose two reduction techniques by exploiting spatial and temporal redundancy in pedestrian videos.
Table~\ref{tab:ablation1} shows that our spatial reduction strategy cuts down the FLOP to approximately the same level as baseline networks while only incurring $0.3\%$ R1/mAP drop on MARS and $0.2\%$ mAP drop on DukeV.
%
%
%
As for temporal reduction, we save $32.6\%$ of FLOP from NVAN and sustain only $1.1\%$ R1 loss on both datasets and $1.7\%$ and $1.2\%$ mAP loss.
%
%
Finally, by applying both spatial and temporal reduction techniques on NVAN, which is our STE-NVAN, we achieve $\mathbf{72.7\%}$ FLOP reduction compare to NVAN and requires $\mathbf{45.7\%}$ less FLOP compare to the baseline that doesn't employ any attention mechanism.
It shows that our proposed STE-NVAN not only improves the Re-ID performance but also demonstrates a more efficient method of extracting sequence features.

%
%
%
\vspace{-3mm}
\paragraph{Analysis of NVAN.}
To better understand the property of non-local operations, we conduct analysis on NVAN regarding RRS strategy and number of inserted non-local attention layers.
%
In Table~\ref{tab:T}, we discover that by increasing the number of frames $T$ sampled from a sequence in RRS, Re-ID performance increases steadily as more frames provide richer information about a pedestrian.
We pick $T=8$ for our NVAN and STE-NVAN in consideration of the memory capacity of our machine.
%
On the other hand, we observe performance gain as we insert more non-local attention layers.
In Table~\ref{tab:non}, we insert a non-local layer at $conv4\_6$ for ``1 layer'' and insert 3 non-local layers at $conv3\_4, conv4\_5, conv4\_6$ for ``3 layers''.
%
%
We insert 5 non-local layers for NVAN and STE-NVAN since it performs the best.

\vspace{-3mm}

\paragraph{Analysis of STE-NVAN.}
Next we investigate the parameters for designing STE-NVAN.
Starting from NVAN, we apply the spatial reduction techniques to group features into horizontal stripes in non-local attention layer.
Table~\ref{tab:n_stripe} shows that while increasing number of stripes $S$ does not introduce excessive additional FLOP, it improves the Re-ID performance subtly.
%
%
As for analyzing temporal reduction, we increase the pooling operations throughout the network.
For comparison, ``in 3DAP'' in Table~\ref{tab:pool_pos} is the NVAN model that pools all features after the last convolutional layer.
By employing additional pooling after the non-local layers located in $stage 4$ (``+ stage 4''), we reduce $10.7\%$ of FLOP from NVAN.
And by introducing another additional pooling after non-local layers at $stage 3$ (``+ stage 3''), we remove $32.7\%$ of FLOP from NVAN while only dropping $0.8\%$ and $0.7\%$ of R1 on MARS and DukeV.
%
%
\begin{table}[t]
    \begin{minipage}{.5\linewidth}
      \centering
      \caption{Comparison of NVAN network with different \# frames of RRS strategy.}
        \begin{tabular}{c|cc|cc}
        \hline
        \multirow{2}{*}{\# frames}  & \multicolumn{2}{c|}{MARS} & \multicolumn{2}{c}{DukeV} \\
        \cline{2-5}
        & R1  & mAP & R1  & mAP      \\ 
        \hline \hline
        $T=4$    & 89.0       & 81.0      & 95.3     &  92.7    \\
        $T=6$    &  89.4      & 81.6      & 95.6  & 93.4 \\ 
        $T=8$   & 90.0      & 82.8      & 96.3       & 94.9  \\
        \hline
     \end{tabular}
    \label{tab:T}
    \caption{Comparison of NVAN network with different \# non-local layers embedded.}
        \begin{tabular}{c|cc|cc}
        \hline
        \# non-local   & \multicolumn{2}{c|}{MARS} & \multicolumn{2}{c}{DukeV} \\
        \cline{2-5}
        layers & R1  & mAP & R1  & mAP      \\ 
        \hline \hline
        $1$ layer     & 89.0       & 81.8     &  95.8    & 93.7     \\
        $3$ layers    & 89.0      &  82.4      & 96.3  &  94.9\\ 
        $5$ layers   & 90.0      & 82.8      & 96.3      & 94.9  \\ 
        \hline
     \end{tabular}
    \label{tab:non}
    \end{minipage}%
    \hspace{0.5mm}
    \begin{minipage}{.5\linewidth}
     \centering
      \caption{Comparison of different \# stripes in spatial reduction non-local layer.}
        \begin{tabular}{c|c|c|c}
        \hline
        \multirow{2}{*}{\# stripes}  & MARS & DukeV & \multirow{2}{*}{\#FLOP}  \\
        \cline{2-3}
        & R1  & R1 &     \\ 
        \hline \hline
        $S=4$        & 89.6    &  96.3  & 30.4G \\
        $S=8$        & 89.5    &  96.1  & 30.4G\\ 
        $S=16$         & 89.7      & 96.3 & 30.4G\\
        \hline
        \end{tabular}
        \label{tab:n_stripe}
        \caption{Comparison of different pooling position combinations in hierarchical structure.}
        \begin{tabular}{c|c|c|c}
        \hline
        Pooling & MARS & DukeV&\multirow{2}{*}{\#FLOP}\\
        \cline{2-3}
        positions & R1  & R1  &       \\ 
        \hline \hline
        in 3DAP    & 90.0          & 96.3      &  60.0G  \\
        $+$stage 4   &  89.8      & 96.1 &  53.6G \\ 
        $+$stage 3   &  89.2      &  95.6    &   40.4G \\ 
        \hline
        \end{tabular}
        \label{tab:pool_pos}
    \end{minipage} 
    \vspace{-5mm}
\end{table}

\subsection{Comparison with State-of-the-arts Approaches}
Table~\ref{tab:sota} reports the comparison of our NVAN and STE-NVAN to state-of-the-art video-based person Re-ID approaches. 
For STA~\cite{sta}, we display their results sampling 8 images per sequence to be fair with our method.
On MARS, our NVAN achieves $\mathbf{90.0\%}$ in R1 and $\mathbf{82.8\%}$ in mAP, surpassing all methods by a large margin.
Our efficient STE-NVAN also performs better than all methods in R1 and breaks even with STA in mAP despite using less FLOP than NVAN.
%
%
On the other hand, our NVAN and STE-NVAN still displays competitive results on DukeV, where Re-ID on DukeV is easier than MARS since detection are manually annotated.
The superior Re-ID performance on two benchmark datasets proves the value of applying non-local operations for extracting a better representation of videos. 
%
%
%
%

\begin{figure*}[t]
	\centering
    \includegraphics[width=0.7\textwidth]{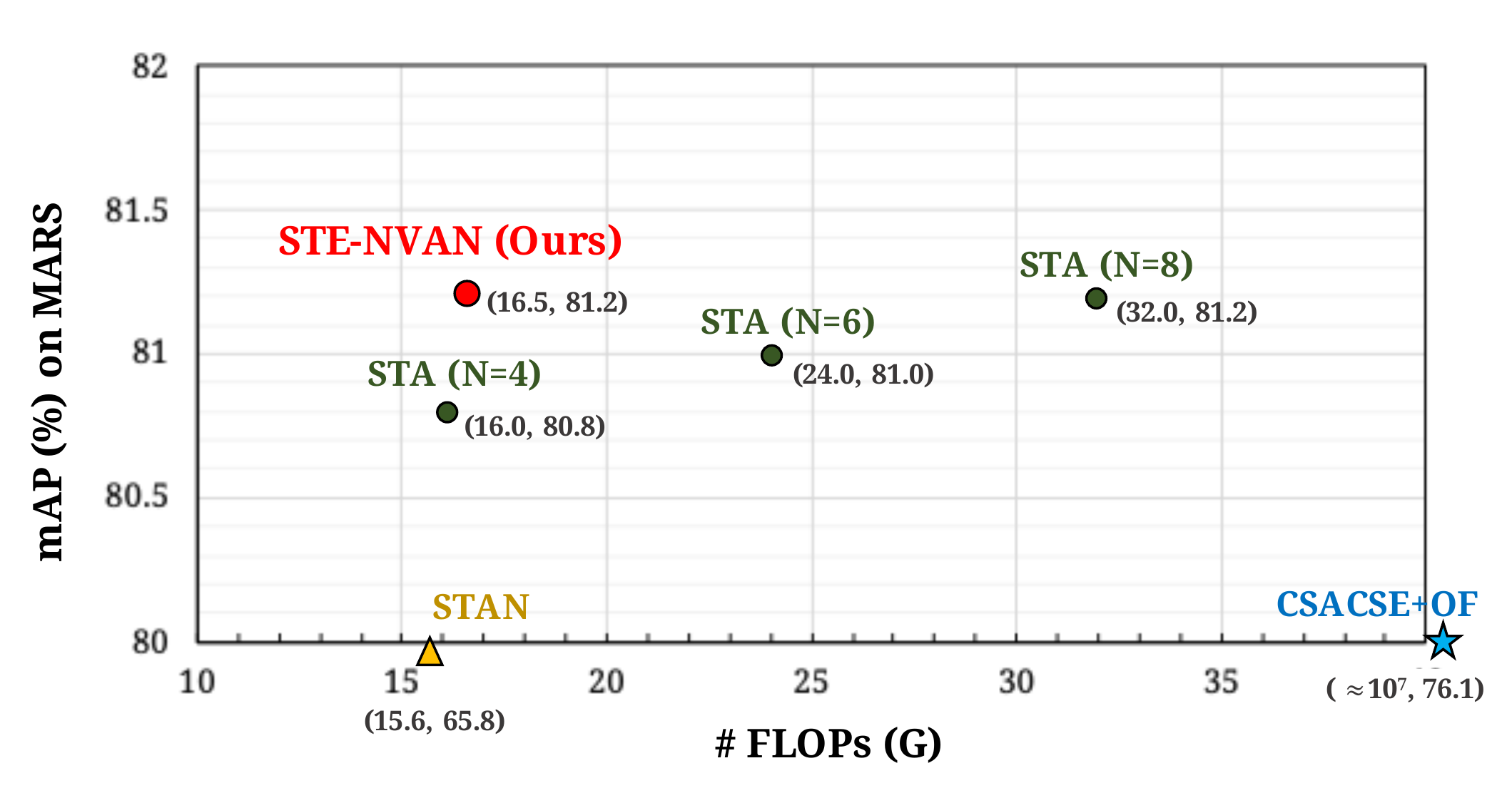}
    \mycaption{Computation-performance plot of our proposed STE-NVAN and existing methods with attention mechanisms}{}
    \label{fig:trade}
    \vspace{-3mm}
\end{figure*}

To take the computation complexity into consideration, we compare our method with existing methods that also uses attention mechanisms on the performance-computation plot in Figure~\ref{fig:trade}. We visualize mAP on MARS dataset for the performance and \#FLOP for computation counts.
%
For STA, we report three variants of their with different numbers of sampled frames per sequence to better demonstrate their trade-off.
Results show that our proposed STE-NVAN exhibits a much better mAP-FLOP trade-off compared to current state-of-the-arts.
STAN~\cite{diversity} and CSACSE+OF~\cite{snip} even lands outside of the plot since their mAP and FLOP are beyond the scale of our plot.
The results not only indicates the advantage of our proposed spatial and temporal reduction techniques but also reveal the importance of considering computation complexity when design feature extractors for video sequences.
%
%
%
%
%

\begin{table}[t]
    \centering
    \mycaption{Comparison with state-of-the-arts approaches on MARS and DukeV}{}
    \label{tab:sota}
   \begin{tabular}{l|c|cc|cc}
        \hline
       \multirow{2}{*}{Methods} & \multirow{2}{*}{Source} & \multicolumn{2}{c|}{MARS} & \multicolumn{2}{c}{DukeV}\\
       \cline{3-6}
        & &R1  & mAP   &R1  & mAP   \\ 
        \hline \hline
        CNN+Kiss.~\cite{mars}    & ECCV16       & 65.0     & 45.6    & - & -  \\
        SeeForest~\cite{forest}  & CVPR17       & 70.6     & 50.7    & - & - \\
        LatentParts~\cite{latent_part} & CVPR17 & 70.6     & 50.7    & - & -  \\
        TriNet~\cite{trip} & arXiv17            & 79.8     & 67.7    & - & -  \\
        ETAP-Net(supervised)~\cite{etap}&CVPR18 & 80.8     & 67.4    & 83.6 & 78.3 \\
        STAN~\cite{diversity} & CVPR18          & 82.3     & 65.8    & -    &  -\\
        CSACSE+OF~\cite{snip}  &  CVPR18        & 86.3     & 76.1    & -    &  -\\
        STA (N=8)~\cite{sta}  & AAAI19                & 86.2     & 81.2    & 96.0 &  \textbf{95.0} \\
        \hline
        NVAN (ours)    & -      & \textbf{90.0} & \textbf{82.8}& \textbf{96.3} & 94.9 \\
        STE-NVAN (ours)   & -       & 88.9      &   81.2    & 95.2  & 93.5\\
        \hline
     \end{tabular}
\vspace{-3mm}
\end{table}

\vspace{-5mm}
\section{Conclusion}
\vspace{-3mm}
We introduce a Non-local Video Attention Network (NVAN) which incorporates multiple non-local attention layers to extract spatial and temporal video characteristics from low to high feature levels, which enrich the representation of videos in person re-identification. To alleviate the computation cost, we proposed a Spatially and Temporally Efficient Non-local Video Attention Network (STE-NVAN), which spatially reduce the non-local operation by utilizing pedestrian part characteristics and temporally reduce the operation with hierarchical structure. Extensive experiments are conducted to prove that our STE-NVAN is a superior trade-off between performance and computation.
\section*{Acknowledgment}
This research was supported in part by the Ministry of Science and Technology of Taiwan (MOST 108-2633-E-002-001), National Taiwan University(NTU-108L104039), Intel Corporation, Delta Electronics and Compal Electronics.
\bibliography{bmvc_ref}
\end{document}